\documentclass[conference]{IEEEtran}
\IEEEoverridecommandlockouts

\usepackage{amsmath,amsfonts}
\usepackage{array}
\usepackage{textcomp}
\usepackage{url}
\usepackage{hyperref}
\usepackage{verbatim}
\usepackage{graphicx}
\usepackage{xcolor}
\usepackage{multirow}
\usepackage{graphicx,xfp}
\usepackage{booktabs,threeparttable}

\usepackage{caption}
\usepackage{subcaption}

\usepackage{pgf-pie} 

\usepackage{tikz}
\usetikzlibrary{positioning}
\usepackage{pgfplots}
\usepackage{marvosym}

\usepackage{tikz}
\usepackage{tikzscale}
\usepackage{xspace}
\usepackage{pgfplots}
\usepgfplotslibrary{groupplots}
\pgfplotsset{compat=1.16}
\usepackage{xparse}
\NewDocumentCommand{\Log}{o}{%
  \IfNoValueTF{#1}{}{{}^{#1}\!}\log}%

\AtBeginDocument{%
  }

\begin{document}

\title{Information We Can Extract About a User From ‘One Minute Mobile Application Usage'}

\author{\IEEEauthorblockN{Sarwan Ali}
\IEEEauthorblockA{\textit{Department of Computer Science} \\
\textit{Georgia State University}\\
Atlanta, GA, USA \\
sali85@student.gsu.edu}
}

\maketitle
\begin{abstract}
Understanding human behavior is an important task and has applications in many domains such as targeted advertisement, health analytics, security, and entertainment, etc. For this purpose, designing a system for activity recognition (AR) is important. However, since every human can have different behaviors, understanding and analyzing common patterns become a challenging task. Since smartphones are easily available to every human being in the modern world, using them to track the human activities becomes possible. In this paper, we extracted different human activities using accelerometer, magnetometer, and gyroscope sensors of android smartphones by building an android mobile applications. Using different social media applications, such as Facebook, Instagram, Whatsapp, and Twitter, we extracted the raw sensor values along with the attributes of $29$ subjects along with their attributes (class labels) such as age, gender, and left/right/both hands application usage. We extract features from the raw signals and use them to perform classification using different machine learning (ML) algorithms. Using statistical analysis, we show the importance of different features towards the prediction of class labels. In the end, we use the trained ML model on our data to extract unknown features from a well known activity recognition data from UCI repository, which highlights the potential of privacy breach using ML models. This security analysis could help researchers in future to take appropriate steps to preserve the privacy of human subjects.
\end{abstract}




\maketitle

\section{Introduction}
Nowadays, our mobile phones have evolved into powerful handheld computing devices. Holding a smartphone is now part of our daily activities. This is why many research studies explored smartphone sensors such as an accelerometer and gyroscope to detect human activities~\cite{shoaib2014fusion}. In general,  one of the most significant tasks in pervasive computing is to provide precise and timely information about people's activities and behaviors~\cite{lara2012survey}. In fact, the study of human activities has emerged as one of the most active research areas in computer vision~\cite{vrigkas2015review}.

Authors in~\cite{kim2009human} discussed the following four challenges that people might face when studying human activities. First, human activities are concurrent; different actions might be happening at the same time. Secondly, human activities are intertwined. Thirdly, similar activities can be  perceived in various ways. Finally, there might be more than one user in the studied environment. All of these human factors alongside technological challenges can contribute to the difficulties of accurately predicting human activities. They also differentiated between ‘Activity Recognition’ and ‘Activity Pattern Discovery’. They defined ‘Activity Recognition’ as the detection of human activities accurately using a predetermined activity model, which means that the model is built first then it gets implemented into a suitable pervasive system. On the other hand, ‘Activity Pattern Discovery’ means that the data is collected first and then models are applied to discover some information and unknown patterns about the activities, which means that the pervasive system is built first and then the sensor data gets analyzed to uncover some patterns.

In this paper, we adopt the ‘\textit{Activity Pattern Discovery}’ method and collect data from lower level smartphone sensors to reveal some information about the user. First, we want to observe whether we can predict what type of mobile (social media) application the user is currently running. We chose the following four applications for our experiments : (Facebook, Instagram, Whatsapp, Twitter). Then, after getting the sensor patterns (using different built-in sensors in the smart device), we want to analyse them further and identify hidden patterns (if there is any such pattern) to uncover additional information about the experiment subjects like : what is the gender of the subject? Do females have different behavior patterns than males? What age group the person belongs to? Can we predict the age of the subject while using some mobile application? And finally we want to observe whether we can recognize if the user is using his left hand, right hand or both hands while browsing the social media application. All of this information might help in recognizing if different users have different experiences while using any application and whether certain patterns disclose some private information about the smartphone users. In the real-world, such a system could be used to enhance user experience while designing any mobile application, for getting likes/dislikes of people, for targeted advertisement, for fall detection (specifically for older people), and for securing the private information of the users based on the detected patterns from the sensor data.

Using the existing data (of $60$ seconds social media application usage), we show that the private information of users can be predicted using simple machine learning models. This behavior shows the vulnerability of the currently available activity data on the internet. Using the analysis we show in this paper, relevant authorities could take appropriate steps to ensure the privacy of the data of people so that it can be avoided to be misused.

In this paper, our contributions are the following:
\begin{enumerate}
    \item We collect human activity data using three sensors of the android smartphones namely accelerometer, magnetometer, and gyroscope by building an android application.
    \item By using our application (for data collection) for just $60$ seconds, we were able to classify different attributes related to human activities with high accuracy.
    \item Using statistical analysis (information gain), we show that some features are important for the classification of certain classes. 
    \item We show that privacy of the existing data could be exploited very easily, which requires relevant authorities to take appropriate steps to avoid the misuse of activity information.
\end{enumerate}

The rest of the paper is organised as follows: Section~\ref{sec_related_work} contains the related work for the human activity recognition problem. Our main methodology is introduced in Section~\ref{sec_Methodology}. Details related to the experiments and dataset statistics are given in Section~\ref{sec_exp_setup}. Our results are reported in Section~\ref{sec_results_discussion}. Finally, we conclude the paper in Section~\ref{sec_conclusion}

\section{Related Work}\label{sec_related_work}
Human activity recognition (HAR) is a well studied problem in the literature~\cite{kaghyan2012activity,derawi2013gait,polu2018human,al2016activity}. Authors in~\cite{kaghyan2012activity} use K-nearest neighbor algorithm for HAR. Using machine learning based algorithms for HAR is done in~\cite{polu2018human}. Authors in~\cite{scholl2012feasibility} used accelerometer data from wrist watches to detect different behaviors of humans. Similarly, authors in~\cite{varkey2012human} use data from wrist watches to classify different human exercises using the ML models.

Converting the unstructured data into a fixed-length numerical representation is an interesting and ongoing research problem. It has been studied in many fields such as
graphs~\cite{hassan2020estimating,Hassan2021Computing}, nodes in
graphs~\cite{ali2021predicting}, and electricity
consumption~\cite{ali2019short,Ali2020ShortTerm}, texts analytics~\cite{Shakeel2020LanguageIndependent,Shakeel2020Multi,Shakeel2019MultiBilingual},
electroencephalography and electromyography
sequences~\cite{ullah2020effect},
Networks~\cite{Ali2019Detecting}, and biological
sequences~\cite{leslie2002mismatch,farhan2017efficient,Kuksa_SequenceKernel,ali2021effective,ali2021simpler,ali2021spike2vec}. For time series data, several authors proposed machine learning and deep learning based methods for time series classification~\cite{ismail2020inceptiontime,hochreiter1997long,wang2017time,chung2014empirical}. However, since deep learning based models are ``data hungry", they cannot be applied when we have limited data~\cite{marcus2018deep}.

\section{Proposed Method}\label{sec_Methodology}
In this section, we discuss the proposed approach in detail, which includes data collection and generating fixed-length numerical representation from the raw sensor data.

\subsection{Mobile Application Usage}
In the first step, we designed an android mobile application that takes some basic information (we use this information as class labels later on) from the user such as the type of social media application they want to use, their gender, age, and the hand in which they hold the smart phone while browsing the social media applications (this information is going to help us in training our classification algorithms). When they click on the start button, the application starts collecting the information from three different sensors namely accelerometer, magnetometer, and gyroscope in the background. The user minimizer our application and start using the required social media application for $60$ seconds. The social media application, which user can use are Facebook, Instagram, Whatsapp, Twitter. After $60$ seconds, the application stops collecting the data and sends a message to the user that the data collection process is stopped. The final data for each sensor is added to separate excel sheets located at the download folder of the mobile. The user then sends the data to us for further analysis.

\subsection{Data Processing and Feature Extraction}
Once the data are received, we then start extracting the features.
Because the raw sensor data can be of varying lengths, and the machine learning (ML) algorithms require a fixed-length vector as input, we did not use the raw data in the analysis and chose specific features only to get a fixed-length feature vector.  Although there are methods that could be used to make the (raw) signal of a fixed-length (such as data padding). However, this approach will only increase the dimensionality of the data (will cause a curse of dimensionality) and hence the classifiers' runtime. Furthermore,  from the information gain analysis (explained in the later section), we discovered that only a few features are important to predict different labels from our data. This means that analysing the raw signal would not help much in improving the classifier's performance and would have only added a huge computational cost. The features selected are Mean, Median, Mode, Quantiles (Divide data into 3 intervals with equal probability), Population Standard Deviation, Sample Standard Deviation, and Variance (total $9$ features). Several ways has been proposed in the literature to use combination of different statistical features to get the fixed-length representation from the raw signals~\cite{ullah2020effect}. However, there is no standard way defined to select a specific types of features. Therefore, we selected the standard features used for statistical analysis.
In this way, we got $9$ features for x, y, and z-axis each from different sensors. Therefore, there will be a total $9+9+9 = 27$ features for each sensor data. In total we had $81$ Features that include $27$ Magnetometer, $27$ Accelerometer, and $27$ Gyroscope features. The total entries (data samples) that we collected from all $29$ users are $112$, which contains samples for different user using different social media applications (with each sample having $81$ features). 

As for choosing the sensors for the data collection, authors in~\cite{shoaib2014fusion} state that in the field of activity recognition, the accelerometer sensor has acquired the greatest attention in research. They also stated that when aiming to  enhance and boost the performance of activity recognition tasks, the  gyroscope and magnetometer have been integrated alongside the accelerometer sensor. In addition, most of the old android devices only had these three sensors. Therefore, we decided to use these basic sensors in order to reach a larger audience for the data collecting.

\subsection{Classification Step}
\begin{figure*}[h!]
    \centering
    \includegraphics[scale = 0.4] {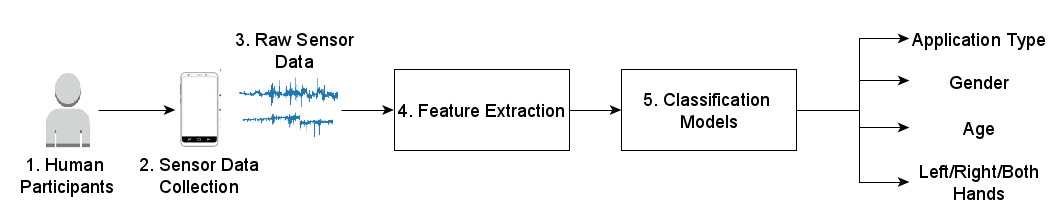}
    \caption{Overview of the system components.}
    \label{fig_flow_chart}
\end{figure*}

After the feature extraction step, we run different machine learning (ML) algorithms to help us correctly identify the user’s behaviour during the mobile application usage. 
We ran seven classification algorithms: Support vector machines(SVM) with a linear kernel, Naive Bayes (NB), Multilayer perceptron (MLP) where we had 10 hidden layers, K-nearest neighbors (KNN) where K=5, Random Forest (RF) with 100 as the number of trees, Logistic regression (LR) with a ‘liblinear’ solver because it is the best choice for  small datasets,  and Decision tree (DT). 
We aimed to compare and contrast the performance of each algorithm in hope that we find one that will  outperform the others. 
In Figure~\ref{fig_flow_chart}, we provide an overview of our system components working together to achieve the goal of predicting different class labels.


\section{Experimental Setup}\label{sec_exp_setup}
In this section, we describe the implementation detail and dataset statistics used for the experimentation.
For mobile application development in android studio (using the java code), we simply used the built-in sensor libraries for Accelerometer, Gyroscope, and Magnetometer. For the ML classification models, we use the python language. Our Java and python code along with the raw and pre-processed data is available online for reproducibility~\footnotetext{Available in the published version}.

We use the standard 5 fold cross validation approach for the experiments and report the average results of $5$ runs. We use different evaluation Metrics to measure the performance of classifiers. The metrics used are : Average Accuracy, Precision, Recall, F1 (weighted), F1 (Macro),F1 (Micro), and ROC AUC (one-vs-rest). 
For computing the information gain (statistical analysis), we use Weka software. 

\subsection{Dataset Statistics}

\begin{figure}
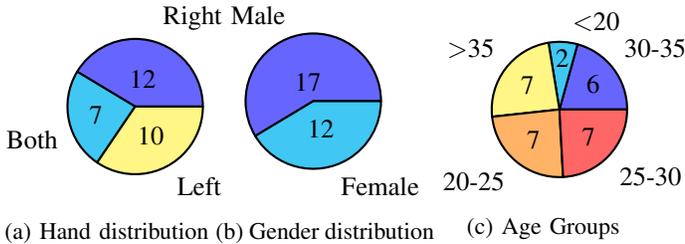

\centering
\begin{subfigure}{.16\textwidth}
    \centering
    \includegraphics{Figures/Tikz/hand_file.tikz}
    \caption{Hand distribution}
\end{subfigure}%
\begin{subfigure}{.16\textwidth}
    \centering
    \includegraphics{Figures/Tikz/gender.tikz}
    \caption{Gender distribution}
\end{subfigure}%
\begin{subfigure}{.16\textwidth}
    \centering
    \includegraphics{Figures/Tikz/age.tikz}
    \caption{Age Groups}
\end{subfigure}%
   \caption{Gender, hand, and age distribution. Figure is best seen in color.}
    \label{fig_data_pie_charts}  
\end{figure}
\begin{figure*}[h!]
\centering
\begin{subfigure}{.33\textwidth}
    \centering
    \includegraphics[scale = 0.13] {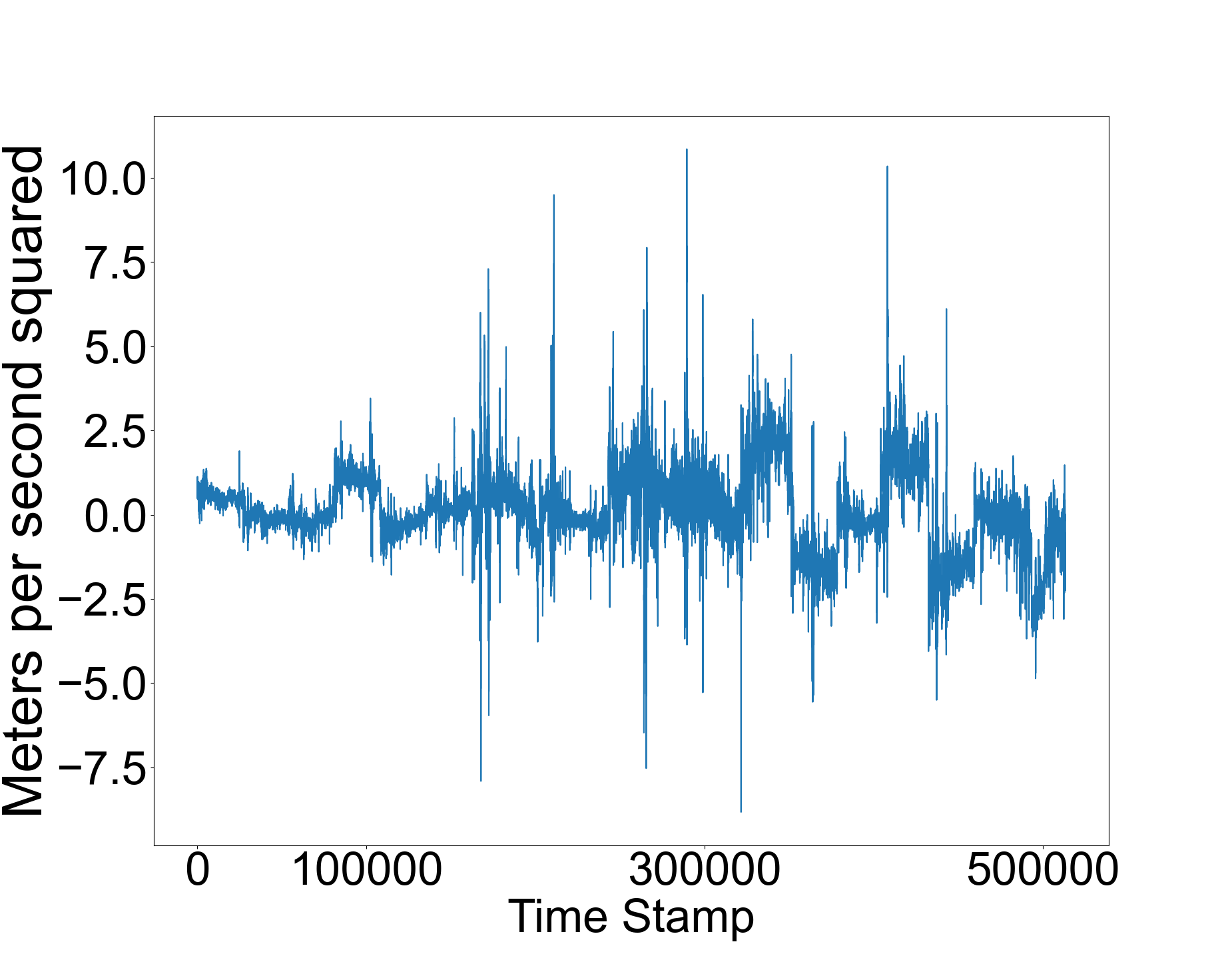}
    \caption{x-axis}
\end{subfigure}%
\begin{subfigure}{.33\textwidth}
    \centering
    \includegraphics[scale = 0.13] {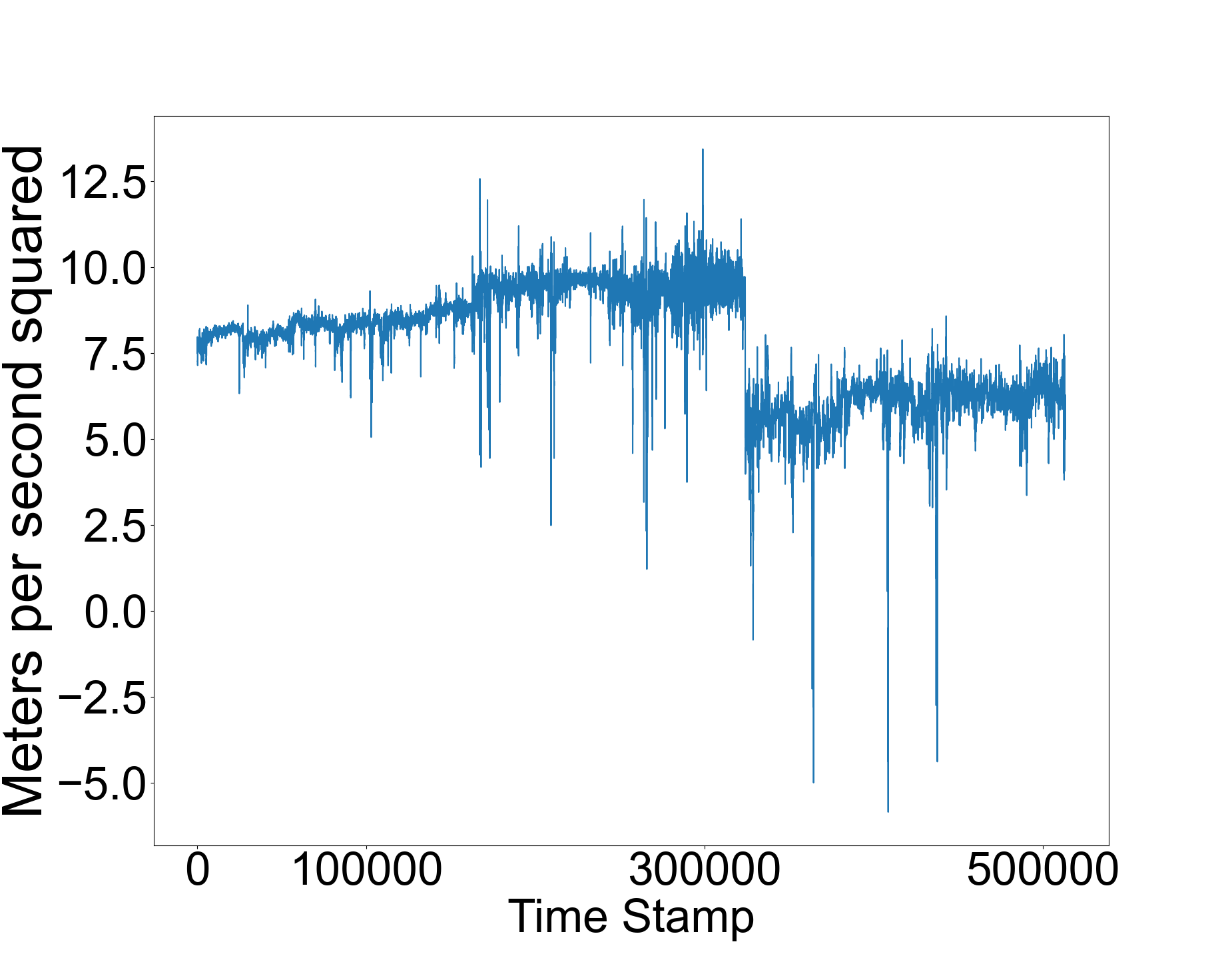}
    \caption{y-axis}
\end{subfigure}%
\begin{subfigure}{.33\textwidth}
    \centering
    \includegraphics[scale = 0.13] {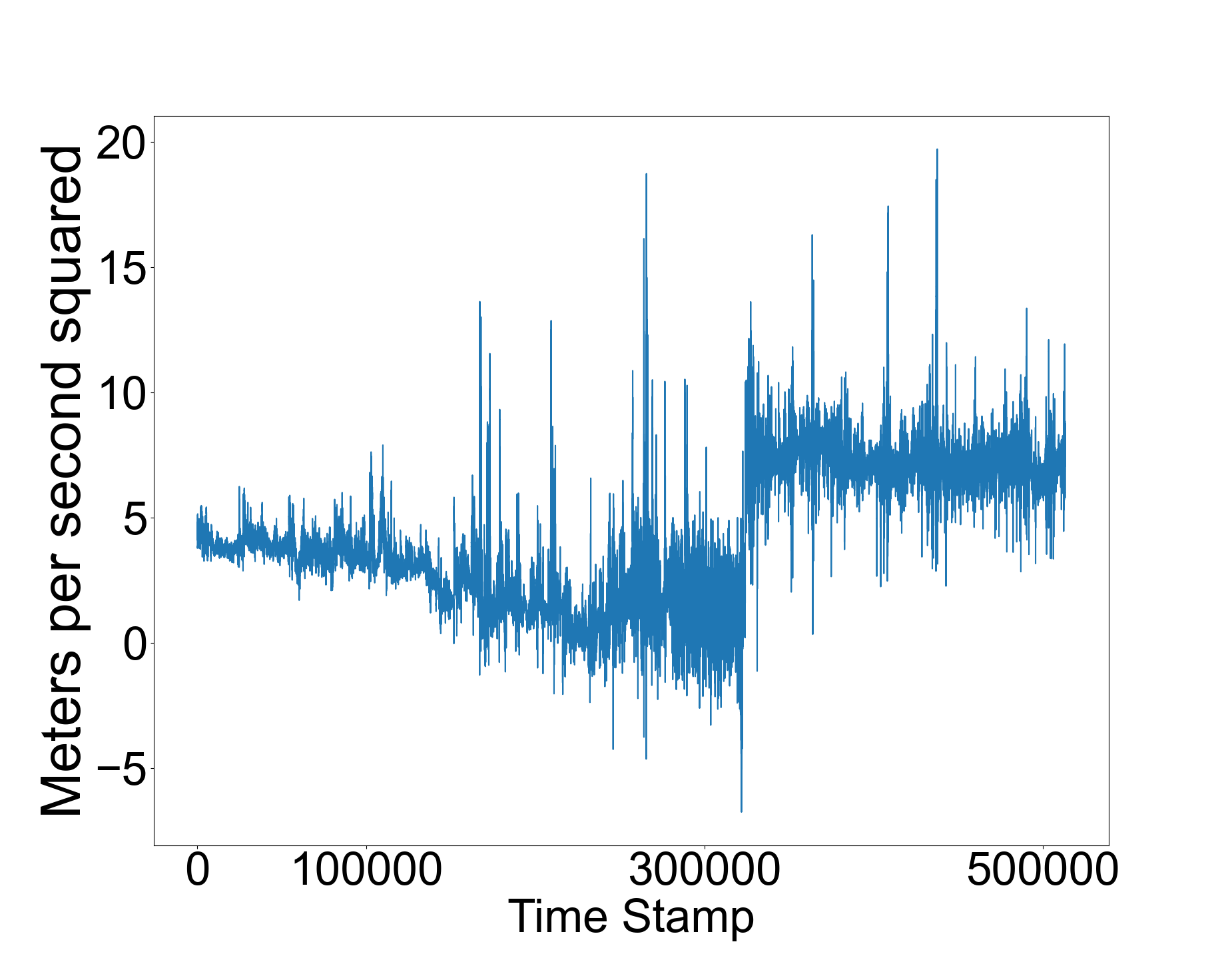}
    \caption{z-axis}
\end{subfigure}%
   \caption{Time series signals for Accelerometer sensor.}
    \label{gif_sensor_signal}   
\end{figure*}

We collected the data from $29$ subjects; $17$ of which were males and the other $12$ were females. Also, $12$ of the $29$ subjects under the study used right handed, $10$ preferred using left handed, while $7$ preferred using mobile applications from both hands. To better illustrate a summary of the collected labels we provide the pie charts in Figure~\ref{fig_data_pie_charts}. Similarly, the signal patterns for different sensors and axis are shown in Figure~\ref{gif_sensor_signal}.

To compute the importance of different features, we use a statistical method, called Information Gain (IG). The IG computes the importance of each feature with respect to the class label. More formally:
    $IG(Class,position) = H(Class) - H(Class | position)$,
where $H=\sum_{ i \in Class} -p_i \log p_i$ is the entropy, and $p_i$ is the
probability of the class $i$.

\section{Results and Discussion}\label{sec_results_discussion}
In this section, we first show the classification results for different class labels. After that, we show the procedure for data prediction for unknown data using our trained models.

\subsection{Gender Prediction}
\begin{figure}[h!]
    \centering
    \includegraphics[scale = 0.9]{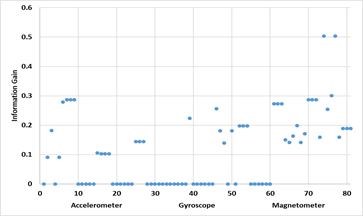}
    \caption{Importance of the features (accelerometer, gyroscope, and magnetometer) for Gender label.}
    \label{fig_gender_label}
\end{figure}

Table~\ref{tbl_gender_results} shows a summary of the ML algorithms for gender prediction (best values are shown in bold). SVM performed the best on this task. Figure~\ref{fig_gender_label} illustrates the information gain values for different features. A feature having higher information gain value means that it is contributing more towards the prediction of the class label (gender). We can observe that large number of features are contributing towards the prediction of gender because of which, we are getting the classification accuracy of $98.2 \%$ (for SVM).

\begin{table}[h!]
  \centering
  \caption{Performance of the ML algorithms on gender prediction. Best values are shown in bold.}
  \resizebox{0.96\linewidth}{!}{
  \begin{tabular}{ccccccc | p{1.1cm}}
    \toprule
    \multirow{3}{0.7cm}{ML Algo.} & \multirow{3}{*}{Acc.} & \multirow{3}{*}{Prec.} & \multirow{3}{*}{Recall} & \multirow{3}{0.9cm}{F1 weigh.} & \multirow{3}{0.9cm}{F1 Macro} & \multirow{3}{1.2cm}{ROC- AUC} & Train. runtime (sec.) \\	
    \midrule	\midrule	
SVM & \textbf{0.982} & \textbf{0.984} & \textbf{0.982} & \textbf{0.982} & \textbf{0.981}  & \textbf{0.980} & 0.011 \\
NB & 0.791 & 0.814 & 0.791 & 0.778 & 0.754  & 0.749 & 0.014 \\
MLP & 0.979 & 0.981 & 0.979 & 0.979 & 0.977  & 0.976 & 0.248 \\
KNN & 0.905 & 0.917 & 0.905 & 0.902 & 0.893 & 0.886 & 0.014 \\
RF & 0.941 & 0.949 & 0.941 & 0.940 & 0.935 & 0.932 & 0.132 \\
LR & 0.976 & 0.979 & 0.976 & 0.976 & 0.974 & 0.972 & 0.013 \\
DT & 0.851 & 0.862 & 0.851 & 0.850 & 0.838 & 0.842 & \textbf{0.008} \\

    \bottomrule
  \end{tabular}
  }
  \label{tbl_gender_results}
\end{table}

\subsubsection{Hand Prediction}
\begin{figure}[h!]
    \centering
    \includegraphics[scale = 0.9]{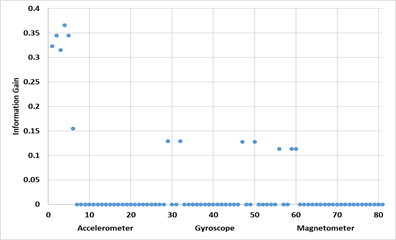}
    \caption{Importance of the features (accelerometer, gyroscope, and magnetometer) for Hand label.}
    \label{fig_hand_label}
\end{figure}

It was really difficult to accurately predict the hand used by the subjects as shown in Table~\ref{tbl_hand_results} (best values are shown in bold). To help us understand why that happens, we used information gain to reveal if there is a correlation between the class label (hand) and the data features. Generally, when information gain value for more features is higher, it implies that those features are (positively) related to the class label, and hence they will contribute more towards increasing the predictive accuracy of the classifiers. However, as shown in Figure~\ref{fig_hand_label}, there are only a few attributes/features with larger information gain. As a result, just a few features actually aid the classifiers in distinguishing between distinct hands. 

\begin{table}[h!]
  \centering
  \caption{Performance of the ML algorithms on hand prediction. Best values are shown in bold.}
  \resizebox{0.96\linewidth}{!}{
  \begin{tabular}{ccccccc | p{1.1cm}}
    \toprule
    \multirow{3}{0.7cm}{ML Algo.} & \multirow{3}{*}{Acc.} & \multirow{3}{*}{Prec.} & \multirow{3}{*}{Recall} & \multirow{3}{0.9cm}{F1 weigh.} & \multirow{3}{0.9cm}{F1 Macro} & \multirow{3}{1.2cm}{ROC- AUC} & Train. runtime (sec.) \\	
    \midrule	\midrule	
SVM & 0.404 & 0.452 & 0.404 & 0.408 & 0.378 & 0.548 & 0.015 \\
NB & 0.404 & \textbf{0.504} & 0.404 & 0.412 & 0.393 & 0.562 & 0.016 \\
MLP & 0.416 & 0.450 & 0.416 & 0.416 & 0.390 & 0.555 & 0.210 \\
KNN & 0.330 & 0.415 & 0.330 & 0.332 & 0.321 & 0.518 & 0.017 \\
RF & 0.410 & 0.440 & 0.410 & 0.405 & 0.374 & 0.547 & 0.140 \\
LR & 0.423 & 0.441 & 0.423 & 0.413 & 0.380 & 0.556 & 0.019 \\
DT & \textbf{0.434} & 0.461 & \textbf{0.434} & \textbf{0.433} & \textbf{0.399} & \textbf{0.563} & \textbf{0.010} \\
    \bottomrule
  \end{tabular}
  }
  \label{tbl_hand_results}
\end{table}

\subsubsection{Application Prediction}
\begin{figure}[h!]
    \centering
    \includegraphics[scale = 0.9]{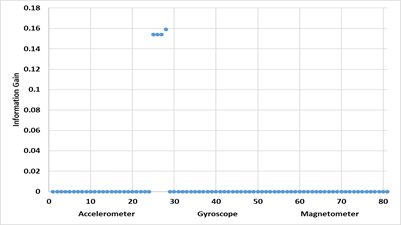}
    \caption{Importance of the features (accelerometer, gyroscope, and magnetometer) for Application label.}
    \label{fig_application_label}
\end{figure}

Table~\ref{tbl_application_results} shows a summary of the results computed using different ML algorithms for social media mobile application prediction (best values are shown in bold). MLP performed the best on this task. We realized  that the mobile applications used in our study (Facebook, Twitter, and Instagram) produce similar clicking patterns. As a result, there are small differences in information gain values for these applications (see Figure~\ref{fig_application_label}). Hence, it is difficult for classifiers to distinguish between applications.

\begin{table}[h!]
  \centering
  \caption{Performance of the ML algorithms on application prediction. Best values are shown in bold.}
  \resizebox{0.96\linewidth}{!}{
  \begin{tabular}{ccccccc | p{1.1cm}}
    \toprule
    \multirow{3}{0.7cm}{ML Algo.} & \multirow{3}{*}{Acc.} & \multirow{3}{*}{Prec.} & \multirow{3}{*}{Recall} & \multirow{3}{0.9cm}{F1 weigh.} & \multirow{3}{0.9cm}{F1 Macro} & \multirow{3}{1.2cm}{ROC- AUC} & Train. runtime (sec.) \\	
    \midrule	\midrule	
SVM & 0.365 & 0.402 & 0.365 & 0.363 & 0.358 & 0.583 & \textbf{0.010} \\
NB & 0.221 & 0.267 & 0.221 & 0.218 & 0.209 & 0.480 & 0.011 \\
MLP & \textbf{0.521} & \textbf{0.574} & \textbf{0.521} & \textbf{0.520} & \textbf{0.515} & \textbf{0.688} & 0.192 \\
KNN & 0.206 & 0.249 & 0.206 & 0.197 & 0.196 & 0.479 & 0.021 \\
RF & 0.477 & 0.542 & 0.477 & 0.478 & 0.473 & 0.660 & 0.132 \\
LR & 0.379 & 0.430 & 0.379 & 0.378 & 0.369 & 0.591 & 0.023 \\
DT & 0.421 & 0.460 & 0.421 & 0.417 & 0.411 & 0.620 & 0.014 \\
    \bottomrule
  \end{tabular}
  }
  \label{tbl_application_results}
\end{table}

\subsubsection{Age Prediction}
\begin{figure}[h!]
    \centering
    \includegraphics[scale = 0.9]{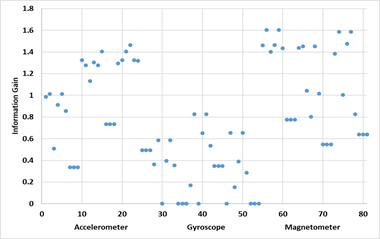}
    \caption{Importance of the features (accelerometer, gyroscope, and magnetometer) for Age label.}
    \label{fig_age_label}
\end{figure}

Table~\ref{tbl_age_results} shows summary of results computed using different ML algorithms for age prediction (best values are shown in bold). MLP performed the best on this task. Figure~\ref{fig_age_label} illustrates different features contribution towards the prediction of age. We can see that since more features have high IG values, therefore, we are getting a higher predictive accuracy as shown in Table~\ref{tbl_age_results}. It means that there is a correlation between the number of features having higher IG value and the classification performance for different ML classifiers.

\begin{table}[h!]
  \centering
  \caption{Performance of the ML algorithms on age prediction. Best values are shown in bold.}
 \resizebox{0.96\linewidth}{!}{
  \begin{tabular}{ccccccc | p{1.1cm}}
    \toprule
    \multirow{3}{0.7cm}{ML Algo.} & \multirow{3}{*}{Acc.} & \multirow{3}{*}{Prec.} & \multirow{3}{*}{Recall} & \multirow{3}{0.9cm}{F1 weigh.} & \multirow{3}{0.9cm}{F1 Macro} & \multirow{3}{1.2cm}{ROC- AUC} & Train. runtime (sec.) \\	
    \midrule	\midrule	
SVM & 0.917 & 0.931 & 0.917 & 0.916 & 0.893 & 0.940 & 0.014 \\
NB & 0.736 & 0.779 & 0.736 & 0.730 & 0.722 & 0.827 & 0.015 \\
MLP & \textbf{0.925} & \textbf{0.938} & \textbf{0.925} & \textbf{0.923} & \textbf{0.905} & \textbf{0.948} & 0.227 \\
KNN & 0.843 & 0.836 & 0.843 & 0.825 & 0.752 & 0.867 & 0.016 \\
RF & 0.915 & 0.924 & 0.915 & 0.911 & 0.884 & 0.934 & 0.139 \\
LR & 0.911 & 0.923 & 0.911 & 0.907 & 0.882 & 0.933 & 0.020 \\
DT & 0.837 & 0.856 & 0.837 & 0.833 & 0.810 & 0.891 & \textbf{0.009} \\

    \bottomrule
  \end{tabular}
  }
  \label{tbl_age_results}
\end{table}

\subsection{UCI Repository Data Analysis}
\begin{figure*}
    \centering
    \begin{subfigure}{.33\textwidth}
        \includegraphics[scale = 1.2]{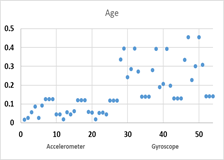}
        \caption{Age label}
    \end{subfigure}%
    \begin{subfigure}{.33\textwidth}
        \includegraphics[scale = 1.2] {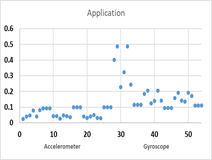}
        \caption{Application label}
    \end{subfigure}%
    \begin{subfigure}{.33\textwidth}
        \includegraphics[scale = 1.2]{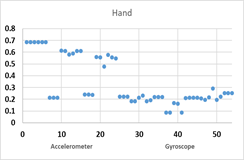}
        \caption{Hand label}
    \end{subfigure}%
    \caption{Importance of the features (accelerometer and gyroscope) for Age, Application, and Hand attributes, respectively (y-axis shows the IG values). We did not show the information gain plot for gender as the best classifier predicted all Males. Since UCI repository data uses only accelerometer and gyroscope sensors, we report the IG values for these two sensors only.}
    \label{fig_feature_importance}
\end{figure*}

To further analyse our results, we use our trained ML models on the UCI Repository data (for testing purpose)~\cite{uci_website} to discover how similar their activity signals (Walking, Walking Upstairs, Walking Downstairs, Sitting, Standing, Laying) are to our collected data. Their dataset had a total of $30$ Subjects. They used the Accelerometer, and Gyroscope sensors to collect the data. They had a Total of $2947$ Entries. Despite the fact that the labeled 'activities' collected in their experiment varied from those collected in ours, it is still possible that the users were walking, laying, sitting, etc. when they were using our mobile application to collect the data. In this case, the collected signals from UCI data and our data should have some correlations. Therefore, we used all of our data, $112$ samples, for training the ML models and all of their data, $2947$ samples of UCI repository data, for testing. Because their labels are different from the ones used in our experiment, the classifiers we trained will predict the labels that it already detected during the training phase (the labels that we have in our data). Due to the label differences, we cannot compute accuracy, precision, etc. for this experiment. Instead, we analyzed the patterns using a contingency table as shown in Table~\ref{tbl_cnt_application}. Table~\ref{tbl_cnt_gender}, Table~\ref{tbl_cnt_age}, and Table~\ref{tbl_cnt_hand} for social media applications, gender, age, and hands (using the best performing classifiers only). We then used the information gain to see the contribution of features towards the prediction of class labels. As previously stated, if more features have high information gain values, the predicted labels are more likely to be correct (because accuracies will be higher as can be seen from Table~\ref{tbl_gender_results} to Table~\ref{tbl_age_results}). Since we can see the higher information gain values in some cases, this indeed means that there is some undiscovered correlation between their data and our data as shown in Figure~\ref{fig_feature_importance} for Age, Application, and Hand labels, respectively (for only the best classifiers from Table~\ref{tbl_hand_results}, Table~\ref{tbl_application_results}, and Table~\ref{tbl_age_results}).

\begin{table}[h!]
  \centering
  \caption{Contingency table for MLP classifier for Application label.}
  \resizebox{0.96\linewidth}{!}{
  \begin{tabular}{ccccc|c}
    \toprule
    Position & Twitter & Facebook & Instagram & Whatsapp & Total \\	
    \midrule	\midrule	
    Laying & 53  & 348  & 32  & 104  & 537  \\
    Sitting & 36  & 326 & 36 & 93 & 491 \\
    Standing & 54 & 321 & 19 & 138 & 532 \\
    Walking & 21 & 169 & 15 & 291 & 496 \\
    Walking Downstairs & 26 & 144 & 11 & 239 & 420 \\
    Walking Upstairs & 26 & 218 & 16 & 211 & 471 \\
    \midrule
    Total & 216 & 1526 & 129 & 1076 & 2947 \\ 
    \bottomrule
  \end{tabular}
  }
  \label{tbl_cnt_application}
\end{table}

\begin{table}[h!]
  \centering
  \caption{Contingency table for SVM classifier for Gender label.}
  \begin{tabular}{cc|c}
    \toprule
    Position & Male & Total \\	
    \midrule	\midrule	
    Laying & 537 &  537  \\
    Sitting & 491 &  491  \\
    Standing & 532 & 532  \\
    Walking & 496 &  496  \\
    Walking Downstairs & 420 &  420  \\
    Walking Upstairs & 471 &  471  \\
    \midrule
    Total & 2947 & 2947 \\ 
    \bottomrule
  \end{tabular}
  \label{tbl_cnt_gender}
\end{table}

\begin{table}[h!]
  \centering
  \caption{Contingency table for MLP classifier for Age label.}
  \resizebox{0.96\linewidth}{!}{
  \begin{tabular}{cccccc|c}
    \toprule
    Position & $<20$ & 20-25 & 25-30 & 30-35 & $>35$ & Total \\	
    \midrule	\midrule	
    Laying & 13 & 140 & 200 & 165 & 19 & 537 \\
    Sitting & 45 & 119 & 121 & 201 & 5 & 491  \\
    Standing & 41 & 123 & 186 & 159 & 23 & 532 \\
    Walking & 1 & 175 & 218 & 31 & 71 & 496 \\
    Walking Downstairs & 2 & 155 & 119 & 71 & 73 & 420 \\
    Walking Upstairs & 3 & 151 & 189 & 80 & 48 & 471 \\
    \midrule
    Total & 105 & 863 & 1033 & 707 & 239 & 2947 \\ 
    \bottomrule
  \end{tabular}
  }
  \label{tbl_cnt_age}
\end{table}

\begin{table}[h!]
  \centering
  \caption{Contingency table for DT classifier for Hand label.}
  \resizebox{0.96\linewidth}{!}{
  \begin{tabular}{cccc|c}
    \toprule
    Position & Left Hand & Right Hand & Both Hands &  Total \\	
    \midrule	\midrule	
    Laying & - & 280 & 257 & 537  \\
    Sitting & 491 & - & - & 491  \\
    Standing & 532 & - & - & 532  \\
    Walking & 496 & - & - &  496 \\
    Walking Downstairs & 420 & - & - &  420 \\
    Walking Upstairs & 471 & - & - &  471 \\
    \midrule
    Total & 2410 & 280  & 257 & 2947 \\ 
    \bottomrule
  \end{tabular}
  }
  \label{tbl_cnt_hand}
\end{table}

\subsection{t-distributed stochastic neighborhood embedding (t-SNE)}
\begin{figure*}
    \centering
    \begin{subfigure}{.5\textwidth}
    \centering
        \includegraphics[scale=1.3] {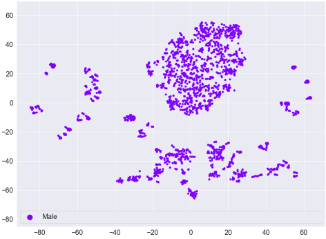}
        \caption{Gender Label}
    \end{subfigure}%
    \begin{subfigure}{.5\textwidth}
    \centering
        \includegraphics[scale=1.3]{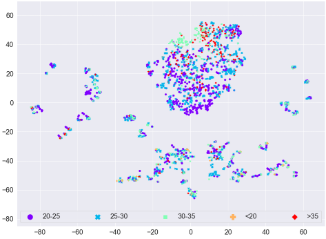}
        \caption{Age Label}
    \end{subfigure}%
    \\
    \begin{subfigure}{.5\textwidth}
    \centering
        \includegraphics[scale=1.3]{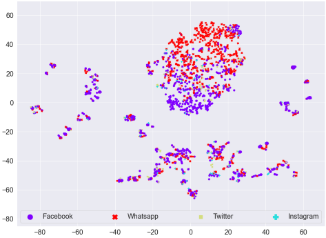}
        \caption{Application Label}
    \end{subfigure}%
    \begin{subfigure}{.5\textwidth}
    \centering
        \includegraphics[scale=1.3]{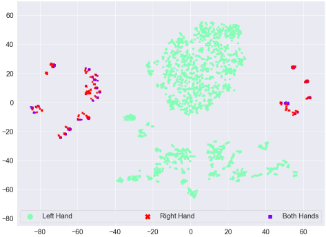}
        \caption{Hand Label}
        \label{fig_uci_tsne_hand}
    \end{subfigure}%
    \caption{t-SNE plot for UCI data. The colored dots in the plots show the predicted labels from the best ML classifiers (MLP for application and age, SVM for gender, and DT for hand).}
    \label{fig_uci_tsne}
\end{figure*}

The t-SNE~\cite{van2008visualizing} is an approach used to get a 2-dimensional representation of the data while preserving the pairwise distance between the data points. This method is ideal for visual data analysis because it allows us to visually see if the data has any natural clustering. We use t-SNE to get 2-D representation of the UCI data and use the predicted labels (from our trained classifiers) to represent the data points by different colors. We can see that there are some patterns in the data, e.g., in Figure~\ref{fig_uci_tsne_hand}, we can notice that those who prefer to use the phone with their left hand (green dots) form separate clusters and are different from people who prefer to use the phone with their right hand or using both hands. The contingency table (in Table~\ref{tbl_cnt_hand}), can also be used to validate this behavior. This behavior shows that most of the subjects involved in collecting the UCI data were left handed, an information which is originally not provided in that data repository.

\section{Conclusion}\label{sec_conclusion}
In summary, we show that using statistical features and standard machine learning models, we can predict the patterns of people, which can be used by researchers and relevant authorities to understand their behavior. Moreover, we show that using information gain, we can observe some correlation between the features in the data and the predictive performance of the ML models. Using this correlation based analysis, we can predict some missing attributes about user, which were not known to the ML models initially. This could be a potential privacy concern because of which, authorities need to take more steps to ensure the privacy of people's data. In future, we will collect more people's data to further analyse the robustness of ML models. Collecting data for applications other than social media could also help us to understand some more hidden features about humans. Similarly, using deep learning for human activity recognition could also be a potential future work.

\bibliographystyle{IEEEtran}
\bibliography{references}

\begin{thebibliography}{10}
\providecommand{\url}[1]{#1}
\csname url@samestyle\endcsname
\providecommand{\newblock}{\relax}
\providecommand{\bibinfo}[2]{#2}
\providecommand{\BIBentrySTDinterwordspacing}{\spaceskip=0pt\relax}
\providecommand{\BIBentryALTinterwordstretchfactor}{4}
\providecommand{\BIBentryALTinterwordspacing}{\spaceskip=\fontdimen2\font plus
\BIBentryALTinterwordstretchfactor\fontdimen3\font minus
  \fontdimen4\font\relax}
\providecommand{\BIBforeignlanguage}[2]{{%
\expandafter\ifx\csname l@#1\endcsname\relax
\typeout{** WARNING: IEEEtran.bst: No hyphenation pattern has been}%
\typeout{** loaded for the language `#1'. Using the pattern for}%
\typeout{** the default language instead.}%
\else
\language=\csname l@#1\endcsname
\fi
#2}}
\providecommand{\BIBdecl}{\relax}
\BIBdecl

\bibitem{shoaib2014fusion}
M.~Shoaib, S.~Bosch, O.~D. Incel, H.~Scholten, and P.~J. Havinga, ``Fusion of
  smartphone motion sensors for physical activity recognition,''
  \emph{Sensors}, vol.~14, no.~6, pp. 10\,146--10\,176, 2014.

\bibitem{lara2012survey}
O.~D. Lara and M.~A. Labrador, ``A survey on human activity recognition using
  wearable sensors,'' \emph{IEEE communications surveys \& tutorials}, vol.~15,
  no.~3, pp. 1192--1209, 2012.

\bibitem{vrigkas2015review}
M.~Vrigkas, C.~Nikou, and I.~A. Kakadiaris, ``A review of human activity
  recognition methods,'' \emph{Frontiers in Robotics and AI}, vol.~2, p.~28,
  2015.

\bibitem{kim2009human}
E.~Kim, S.~Helal, and D.~Cook, ``Human activity recognition and pattern
  discovery,'' \emph{IEEE pervasive computing}, vol.~9, no.~1, pp. 48--53,
  2009.

\bibitem{kaghyan2012activity}
S.~Kaghyan and H.~Sarukhanyan, ``Activity recognition using k-nearest neighbor
  algorithm on smartphone with tri-axial accelerometer,'' \emph{International
  Journal of Informatics Models and Analysis (IJIMA), ITHEA International
  Scientific Society, Bulgaria}, vol.~1, pp. 146--156, 2012.

\bibitem{derawi2013gait}
M.~Derawi and P.~Bours, ``Gait and activity recognition using commercial
  phones,'' \emph{computers \& security}, vol.~39, pp. 137--144, 2013.

\bibitem{polu2018human}
S.~K. Polu and S.~Polu, ``Human activity recognition on smartphones using
  machine learning algorithms,'' \emph{International Journal for Innovative
  Research in Science \& Technology}, vol.~5, no.~6, pp. 31--37, 2018.

\bibitem{al2016activity}
N.~Al-Naffakh, N.~Clarke, P.~Dowland, and F.~Li, ``Activity recognition using
  wearable computing,'' in \emph{2016 11th International Conference for
  Internet Technology and Secured Transactions (ICITST)}.\hskip 1em plus 0.5em
  minus 0.4em\relax IEEE, 2016, pp. 189--195.

\bibitem{scholl2012feasibility}
P.~M. Scholl and K.~Van~Laerhoven, ``A feasibility study of wrist-worn
  accelerometer based detection of smoking habits,'' in \emph{2012 Sixth
  International Conference on Innovative Mobile and Internet Services in
  Ubiquitous Computing}.\hskip 1em plus 0.5em minus 0.4em\relax IEEE, 2012, pp.
  886--891.

\bibitem{varkey2012human}
J.~P. Varkey, D.~Pompili, and T.~A. Walls, ``Human motion recognition using a
  wireless sensor-based wearable system,'' \emph{Personal and Ubiquitous
  Computing}, vol.~16, no.~7, pp. 897--910, 2012.

\bibitem{hassan2020estimating}
Z.~Hassan, M.~Shabbir, I.~Khan, and W.~Abbas, ``Estimating descriptors for
  large graphs,'' in \emph{Advances in Knowledge Discovery and Data Mining
  (PAKDD)}, 2020, pp. 779--791.

\bibitem{Hassan2021Computing}
Z.~Hassan, I.~Khan, M.~Shabbir, and W.~Abbas, ``Computing graph descriptors on
  edge streams,'' 2021,
  \url{https://www.researchgate.net/publication/353671195_Computing_Graph_Descriptors_on_Edge_Streams}.

\bibitem{ali2021predicting}
S.~Ali, M.~H. Shakeel, I.~Khan, S.~Faizullah, and M.~A. Khan, ``Predicting
  attributes of nodes using network structure,'' \emph{ACM Transactions on
  Intelligent Systems and Technology}, vol.~12, no.~2, pp. 1--23, 2021.

\bibitem{ali2019short}
S.~Ali, H.~Mansoor, N.~Arshad, and I.~Khan, ``Short term load forecasting using
  smart meter data,'' in \emph{International Conference on Future Energy
  Systems}, 2019, pp. 419--421.

\bibitem{Ali2020ShortTerm}
S.~Ali, H.~Mansoor, I.~Khan, N.~Arshad, M.~A. Khan, and S.~Faizullah,
  ``Short-term load forecasting using ami data,'' \emph{arXiv preprint
  arXiv:1912.12479}, 2019.

\bibitem{Shakeel2020LanguageIndependent}
M.~H. Shakeel, S.~Faizullah, T.~Alghamidi, and I.~Khan, ``Language independent
  sentiment analysis,'' in \emph{2019 International Conference on Advances in
  the Emerging Computing Technologies (AECT)}, 2020, pp. 1--5.

\bibitem{Shakeel2020Multi}
M.~H. Shakeel, A.~Karim, and I.~Khan, ``A multi-cascaded model with data
  augmentation for enhanced paraphrase detection in short texts,''
  \emph{Information Processing \& Management}, vol.~57, no.~3, p. 102204, 2020.

\bibitem{Shakeel2019MultiBilingual}
M.~Shakeel., A.~Karim, and I.~Khan, ``A multi-cascaded deep model for bilingual
  sms classification,'' in \emph{International Conference on Neural Information
  Processing (ICONIP)}, 2019, pp. 287--298.

\bibitem{ullah2020effect}
A.~Ullah, S.~Ali, I.~Khan, M.~A. Khan, and S.~Faizullah, ``Effect of analysis
  window and feature selection on classification of hand movements using emg
  signal,'' in \emph{SAI Intelligent Systems Conference (IntelliSys)}, 2020,
  pp. 400--415.

\bibitem{Ali2019Detecting}
S.~Ali, M.~K. Alvi, S.~Faizullah, M.~A. Khan, A.~Alshanqiti, and I.~Khan,
  ``Detecting ddos attack on sdn due to vulnerabilities in openflow,'' in
  \emph{International Conference on Advances in the Emerging Computing
  Technologies (AECT)}.\hskip 1em plus 0.5em minus 0.4em\relax IEEE, 2020, pp.
  1--6.

\bibitem{leslie2002mismatch}
C.~Leslie, E.~Eskin, J.~Weston, and W.~Noble, ``Mismatch string kernels for svm
  protein classification,'' in \emph{Advances in neural information processing
  systems (NeurIPS)}, 2003, pp. 1441--1448.

\bibitem{farhan2017efficient}
M.~Farhan, J.~Tariq, A.~Zaman, M.~Shabbir, and I.~Khan, ``Efficient
  approximation algorithms for strings kernel based sequence classification,''
  in \emph{NeurIPS}, 2017, pp. 6935--6945.

\bibitem{Kuksa_SequenceKernel}
P.~Kuksa, I.~Khan, and V.~Pavlovic, ``Generalized similarity kernels for
  efficient sequence classification,'' in \emph{SIAM International Conference
  on Data Mining (SDM)}, 2012, pp. 873--882.

\bibitem{ali2021effective}
S.~Ali, T.~E. Ali, M.~A. Khan, I.~Khan, and M.~Patterson, ``Effective and
  scalable clustering of sars-cov-2 sequences,'' in \emph{International
  Conference on Big Data Research (ICBDR)}, 2021, pp. 42--49.

\bibitem{ali2021simpler}
S.~Ali, S.~Ciccolella, L.~Lucarella, G.~D. Vedova, and M.~Patterson, ``Simpler
  and faster development of tumor phylogeny pipelines,'' \emph{Journal of
  Computational Biology}, vol.~28, no.~11, pp. 1142--1155, 2021.

\bibitem{ali2021spike2vec}
S.~Ali and M.~Patterson, ``Spike2vec: An efficient and scalable embedding
  approach for covid-19 spike sequences,'' in \emph{IEEE International
  Conference on Big Data (Big Data)}, 2021, pp. 1533--1540.

\bibitem{ismail2020inceptiontime}
H.~Ismail~Fawaz, B.~Lucas, G.~Forestier, C.~Pelletier, D.~F. Schmidt, J.~Weber,
  G.~I. Webb, L.~Idoumghar, P.-A. Muller, and F.~Petitjean, ``Inceptiontime:
  Finding alexnet for time series classification,'' \emph{Data Mining and
  Knowledge Discovery}, vol.~34, no.~6, pp. 1936--1962, 2020.

\bibitem{hochreiter1997long}
S.~Hochreiter and J.~Schmidhuber, ``Long short-term memory,'' \emph{Neural
  computation}, vol.~9, no.~8, pp. 1735--1780, 1997.

\bibitem{wang2017time}
Z.~Wang, W.~Yan, and T.~Oates, ``Time series classification from scratch with
  deep neural networks: A strong baseline,'' in \emph{2017 International joint
  conference on neural networks (IJCNN)}.\hskip 1em plus 0.5em minus
  0.4em\relax IEEE, 2017, pp. 1578--1585.

\bibitem{chung2014empirical}
J.~Chung, C.~Gulcehre, K.~Cho, and Y.~Bengio, ``Empirical evaluation of gated
  recurrent neural networks on sequence modeling,'' \emph{arXiv preprint
  arXiv:1412.3555}, 2014.

\bibitem{marcus2018deep}
G.~Marcus, ``Deep learning: A critical appraisal,'' \emph{arXiv preprint
  arXiv:1801.00631}, 2018.

\bibitem{uci_website}
{UCI Machine Learning Repository: Human Activity Recognition Using Smartphones
  Data Set},
  \url{https://archive.ics.uci.edu/ml/datasets/human+activity+recognition+using+smartphones},
  2021, [Online; accessed 01-March-2022].

\bibitem{van2008visualizing}
L.~Van~der M. and G.~Hinton, ``Visualizing data using t-sne.'' \emph{JMLR},
  vol.~9, no.~11, 2008.

\end{thebibliography}

\end{document}